\documentclass[conference,10pt,final]{IEEEtran}
\DeclareUnicodeCharacter{F0FD}{}

\usepackage{amsmath,amssymb,amsfonts}
\usepackage{algorithmic}
\usepackage{graphicx}
\usepackage{array}
\usepackage{microtype}
\usepackage{svg}
\usepackage{textcomp}
\usepackage{tikz}
\usepackage{xcolor}
\usepackage{multirow}
\usepackage{adjustbox}
\usetikzlibrary{shapes.geometric, arrows, decorations.pathreplacing}
\def\BibTeX{{\rm B\kern-.05em{\sc i\kern-.025em b}\kern-.08em
    T\kern-.1667em\lower.7ex\hbox{E}\kern-.125emX}}

\usepackage[backend=biber,style=ieee]{biblatex}
\begin{document}
\title{Beyond Meme Templates: Limitations of Visual Similarity Measures in Meme Matching\\
\thanks{}
}

\author{\IEEEauthorblockN{Muzhaffar Hazman}
\IEEEauthorblockA{\textit{School of Computer Science} \\
\textit{University of Galway}\\
Galway, Ireland \\
m.hazman1@universityofgalway.ie}
\and
\IEEEauthorblockN{Susan McKeever}
\IEEEauthorblockA{\textit{School of Computer Science} \\
\textit{Technological University Dublin}\\
Dublin, Ireland\\
susan.mckeever@tudublin.ie}
\and
\IEEEauthorblockN{Josephine Griffith}
\IEEEauthorblockA{\textit{School of Computer Science} \\
\textit{University of Galway}\\
Galway, Ireland \\
josephine.griffith@universityofgalway.ie}

}

\maketitle

\begin{abstract}
Internet memes, now a staple of digital communication, play a pivotal role in how users engage within online communities and allow researchers to gain insight into contemporary digital culture. These engaging user-generated content are characterised by their reuse of visual elements also found in other memes. Matching instances of memes via these shared visual elements, called \textit{meme matching}, is the basis of a wealth of meme analysis approaches. However, most existing methods assume that every meme consists of a shared visual background, called a \textit{template}, with some overlaid text, thereby limiting meme matching to comparing the background image alone. Current approaches exclude the many memes that are not template-based and limit the effectiveness of automated meme analysis and would not be effective at linking memes to contemporary web-based meme dictionaries. In this work, we introduce a broader formulation of meme matching that extends beyond template matching. We show that conventional similarity measures, including a novel segment‑wise computation of the similarity measures, excel at matching template-based memes but fall short when applied to non-template-based meme formats. However, the segment-wise approach was found to consistently outperform the whole‑image measures on matching non-template-based memes. Finally, we explore a prompting-based approach using a pretrained Multimodal Large Language Model for meme matching. Our results highlight that accurately matching memes via shared visual elements, not just background templates, remains an open challenge that requires more sophisticated matching techniques.

\end{abstract}


\section{Introduction}

Multimodal Internet memes (more commonly, memes) have become a staple medium of communication in almost all online digital communities. Like various forms of multimodal user-generated content, memes combine visual and textual elements to create creative and visually engaging expressions. Uniquely, memes are distinguished from other forms of digital media by their use of shared visual elements also found in other memes \cite{shifman_2013}, forming groups of related memes that express meaning using these elements. We refer to the task of identifying pairs or groups of memes as being related as meme matching approaches. Meme matching provides context for analysing individual memes \cite{hazman_what,murgas_template} and enables the analysis of memes as networks of interconnected expressions \cite{zannetou_origins,courtois_geneaology}.


Many current meme matching approaches are limited to memes that use shared visual backgrounds (\textbf{meme templates}), where two memes are deemed related only if they share the same visual background while other elements are ignored; we refer to this as \textbf{Template Matching} (\texttt{TM}). These approaches require (1) a visual representation, (2) a quantitative similarity measure, and (3) a threshold that determines whether a meme pair qualifies as a match.
Commonly used similarity measures include the distance between image hashes \cite{zannetou_origins}, cosine similarity between embeddings from a pretrained visual encoder \cite{dubey_sparse}, and the distance between descriptors of visual keypoints \cite{courtois_geneaology}. 

However, the relationship between memes is not only defined based on common templates \cite{shifman_2013}. Hazman et al. defined a meme as ``a user-generated piece of digital content that comprises either visual and/or textual components that can be identified as memetic elements'' \cite{hazman_what}. Building on this, we define \textbf{Memetic Matching} (\texttt{MM}) as the task of determining whether two memes share a visual memetic element. Memetic elements \cite{hazman_what} include characters, standalone images and superimposed foreground elements \cite{shifman_photos,cannizaro_2016}, as well as background templates. Although various visual similarity measures have been evaluated on \texttt{TM} \cite{murgas_template}, we found no similar analysis for the \texttt{MM}. 

Currently, studies on memes, e.g. \cite{zannetou_origins,dubey_sparse,bates_templatememe}, tend to rely on a Template Matching step. Since memetic elements take many forms, these analyses ignore a significant portion of the meme typology. Recently, Hazman et al. found that only 55\% of memes randomly sampled from various meme classification datasets were template-based \cite{hazman_what}. Similarly, knowledge-enriched meme analyses that use \texttt{TM} to retrieve semantic information can only access a subset of any knowledge base which reflect this diversity, such as community-maintained meme dictionaries \cite{bates_templatememe} and knowledge graphs \cite{tommasini_imkg}. Accurately matching memes with relevant knowledge base entries on various memetic elements requires meme matching methods that go beyond \texttt{TM}.


\begin{table}[!b]

        \centering
        \begin{tabular}{  >{\centering\arraybackslash}m{0.05\linewidth}  >{\centering\arraybackslash}m{0.15\linewidth}  >{\raggedright\arraybackslash}m{0.2\linewidth}   >{\centering\arraybackslash}m{0.15\linewidth}>{\centering\arraybackslash}m{0.15\linewidth}} 
            \hline
            & \textbf{Image $A$} & \textbf{Image $B$} & \textbf{Memetic Matching (\texttt{MM})} & \textbf{Template Matching (\texttt{TM})}\\
            \hline
            
            1 &\multicolumn{2}{m{0.45\linewidth}}{\centering \includesvg[width=\linewidth]{images/matches/meme_match_0.svg}}  & \texttt{related} & \texttt{unrelated} \\

            2 &\multicolumn{2}{m{0.45\linewidth}}{\centering \includesvg[width=\linewidth]{images/matches/template_match_0.svg}} & \texttt{related}  & \texttt{related} \\

            3 &\multicolumn{2}{m{0.45\linewidth}}{\centering \includesvg[width=\linewidth]{images/matches/meme_match_3.svg}} & \texttt{related} & \texttt{unrelated} \\

            4 &\multicolumn{2}{m{0.45\linewidth}}{\centering \includesvg[width=\linewidth]{images/matches/meme_match_4.svg}} & \texttt{related} & \texttt{unrelated} \\
            \hline 
        \end{tabular}
        \caption{Sample Image Pairs with class labels under the \texttt{MM} and \texttt{TM} tasks.}
        \label{tab:memetic_matching}

\end{table}
In this work, we propose Memetic Matching as a more comprehensive formulation of meme matching than Template Matching. We compare the performance of various visual similarity measures in tackling the two tasks.
Each task is a binary classification problem to determine if a meme is related to (matches) a given reference image, i.e., a dictionary entry image for \texttt{MM} and a template image for \texttt{TM}. 
 The similarity measures are computed using visual representations that are commonly used in existing \texttt{TM} approaches: visual keypoints, pretrained embeddings, and image hashes. We also propose and evaluate a novel segment‐wise representation of embedding and hash similarities. We also evaluate the performance of a pretrained Multimodal LLM (MLLM) in tackling the Memetic Matching task. To evaluate these methods on tackling \texttt{MM} and \texttt{TM}, we prepare a manually labelled dataset for each task.
 
 Our findings reveal several key insights. First, \texttt{MM} is shown to be distinct from and more challenging than \texttt{TM}; when tackled using the visual similarity measures.
 Second,the two tasks are best approached using different similarity measures. Furthermore, we found that segment-wise similarity measures outperformed their whole-image counterparts for \texttt{MM}. Finally, we show that zero-shot and few-shot prompting of an MLLM yield even poorer results for \texttt{MM}. These findings underscore the greater complexity of Memetic Matching and point to the need for more advanced solutions to reliably link memes to entries in community-maintained meme dictionaries.

\section{Datasets}\label{sec:data}
For the two tasks, Template Matching and Memetic Matching, we create separate datasets, $\mathcal{D}^{\texttt{TM}}$ and $\mathcal{D}^{\texttt{MM}}$, respectively. Each row of these datasets is a tuple:
$ (m,r,\hat{y})  \in \mathcal{D}^{(t)} $
where $m$ is a meme image, $r$ is either a dictionary entry image (for $t = \texttt{MM}$) or a template image (for $t = \texttt{TM}$), and $\hat{y}$ indicates whether the pair is \texttt{related} or \texttt{unrelated} per the definition used for the task $t$.

\subsection{Task 1: Template Matching (\texttt{\textup{TM}})}

As is common amongst existing \texttt{TM} works, e.g. \cite{murgas_template}, a template-based definition for meme matching only considers whether a meme $m$ uses a given template $r^{\texttt{TM}}$ as its background. To prepare $\mathcal{D}^{\texttt{TM}}$, we replicate the data source, collection, and annotation methods reported in \cite{courtois_geneaology,murgas_template}. We collected 2,016 templates listed on IMGFlip.com. Similar to \cite{murgas_template}, we removed duplicate templates by extracting all templates with a ResNext embedding cosine similarity greater than 85\% to another template. These were manually reviewed for groups of duplicates, and all but one duplicate in each group was removed. Cropped versions of templates were also considered duplicates. Our final set of reference images, $R^{\texttt{TM}}$, consists of 1,288  IMGFlip template images for $\mathcal{D}^{\texttt{TM}}$.


IMGFlip, by design, links memes with the template that was used to create them. This allows for convenient extraction of \texttt{related} $(m,r)$ pairs \cite{murgas_template}.
For each template, we collected 13 of its most popular memes to create our set of candidate memes. To maintain a class imbalance similar to that of $\mathcal{D}^{\texttt{MM}}$ (see below), we needed a ratio of 1:4 \texttt{related} to \texttt{unrelated} pairs. We randomly selected a meme linked to each template to create a \texttt{related} meme-template image pair. The remaining memes were used to create \texttt{unrelated} pairs, as follows: To find \texttt{unrelated} memes for each template $r^{\texttt{TM}}$, we selected the four templates most similar to $r^{\texttt{TM}}$ by embedding cosine similarity, $r' \in R'$  where $|R'| = 4$ and $r \not equal r^{\texttt{TM}}$. We then randomly sampled one meme that is linked to each $r'$, creating a set of 4 memes that are \texttt{unrelated} to $r^{\texttt{TM}}$. In total, $\mathcal{D}^{\texttt{TM}}$ consists of 1,288 \texttt{related} and 5,152 \texttt{unrelated} pairs.



\subsection{Task 2: Memetic Matching (\texttt{\textup{MM}})}
To create $\mathcal{D}^{\texttt{MM}}$, we needed reference images $r^{\texttt{MM}}$, each representing a dictionary entry.
For each pair $(m,r^{\texttt{MM}})$, its $\hat{y}$ is \texttt{related} if $m$ and $r$ share a visual element according to the definition given in \cite{hazman_what}. Otherwise, the pair is \texttt{unrelated}.

For our dictionary, we chose KnowYourMeme.com (KYM), one of the largest repositories of crowd-sourced and verified information on memes.KYM has been used as a data source in numerous prior works on meme analysis, e.g. \cite{tommasini_imkg,bates_templatememe}. We
focus on the collection of \textit{Confirmed Memes}, comprising articles with verified crowd-sourced information and examples of popular memes. 

We represent each KYM entry by their \textit{header image}, which had been manually selected by the KYM community, or their editorial team, to represent the meme described in the entry, similar to the \textit{infobox} image of a Wikipedia article. We refer to these images as dictionary images, $r^{\texttt{MM}}$. Our collection of dictionary images, $R^{\texttt{MM}}$, was initially seeded with those previously collected by Bates et al. \cite{bates_templatememe}. To expand our coverage of the KYM dictionary, we supplemented this set by retrieving all accessible KYM entries categorised under the \textit{Confirmed Meme} label, as of December 2024. However, due to limitations of the KYM website interface, we were unable to access the entire dictionary. Instead, we systematically collected header images from all entries visible through each of the distinct browsing views provided by the KYM site. This process yielded a final collection of 7,418 dictionary images, which serve as the reference image set $R^{\texttt{MM}}$ for $\mathcal{D}^{\texttt{MM}}$.

Given that not all KYM entries contain example memes, and to better ground our analysis amongst contemporary meme analysis works, the memes for this task are sourced from meme classification datasets \cite{memo1_report,memo2_report,fersini,harmeme}. However, with 35,023 meme and 7,418 dictionary images, we deemed a set of 260k pairs to be unfeasibly large to annotate. Furthermore, we assume that each meme is only related to a few entries. Taking this into account, it is unlikely that random sampling would present sufficient \texttt{related} $(m,r)$ pairs for annotation.
In order to have a balance between matched and unmatched entries in the dataset, we prioritised meme-dictionary image pairs in descending order of embedding cosine similarity.

For some memes, we did not find a \texttt{related} $r^{\texttt{MM}}$ image. In such cases, we only included it in \texttt{unrelated} pairs. For each meme, we annotate at least three \texttt{unrelated} dictionary entries before proceeding to the next meme. The visual similarity in each pair was only used to prioritise which pairs were to be annotated, but was not considered in the annotation itself. In total, $\mathcal{D}^{\texttt{MM}}$ comrpises 7,163 annotated $(m,r^{\texttt{MM}})$ pairs, 1,786 of which are labelled as \texttt{related}, while the remaining 5,819 were labelled as \texttt{unrelated}.

\subsection{\texttt{\textup{TM}} Versus \texttt{\textup{MM}}}
The primary difference between the two tasks lies in the different annotation criteria. Consider, for example, the image pairs and the annotation with respect to \texttt{MM} and \texttt{TM} in Tab. \ref{tab:memetic_matching}. For each pair, $A$ is the meme image $m$ and $B$ is the reference image $r$. Pair 1 shows images with entirely distinct backgrounds, but the person in 1B is reused in 1A twice: once flipped horizontally and another time edited to look older. Although this pair is deemed \texttt{related} under \texttt{MM}, it is deemed \texttt{unrelated} under \texttt{TM} since the backgrounds are not identical. Pair 2 shows a case where image 2B is used as the background for 2A, i.e., as a template, which is classified as \texttt{related} under both \texttt{MM} and \texttt{TM}. Pair 3 shows the memetic reuse in image 3A of a local element, i.e., the \textit{Scumbag Hat}, classifying it as \texttt{related} under \texttt{MM}. However, since 3A does not use 3B as its background, this pair would be classified as \texttt{unrelated} under \texttt{TM}. Pair 4 shows the entirety of image 4B (a recognised meme template) reused as a \textit{panel segment} in image 4A, which is deemed \texttt{related} under \texttt{MM}. However, since image 4B is not used as the whole background of image 4A, Pair 4 is deemed \texttt{unrelated} under \texttt{TM}.

Beyond differences in annotation criteria, we also chose to use distinct pairs $(m,r)$ for each $\mathcal{D}^{\texttt{TM}}$ and $\mathcal{D}^{\texttt{MM}}$; i.e., for $m^t \in M^{t}$ and $ r^t \in R^{t}$, $M^{\texttt{TM}} \not = M^{\texttt{MM}}$ and $R^{\texttt{TM}} \not = R^{\texttt{MM}}$. Using distinct datasets allows us to evaluate \texttt{TM} in the manner it is typically presented in current works (e.g. \cite{murgas_template}) while posing \texttt{MM} in a realistic use case, i.e., retrieval of KYM dictionary entries. Furthermore, each dataset could not also be annotated for the other task. First, since IMGFlip contain only template-based memes and that templates are also considered memetic elements under \texttt{MM} (see Pair 2 in Tab. \ref{tab:memetic_matching}), reannotating the memes in $\mathcal{D}^{\texttt{TM}}$ for the texttt{MM} task. Second, KYM entries for template-based memes in $\mathcal{D}^{\texttt{MM}}$ typically do not use the full template image as its header image\footnote{see for example: https://knowyourmeme.com/memes/scumbag-hat}. Reannotating $\mathcal{D}^{\texttt{MM}}$ for the \texttt{TM} task would yield very few \texttt{related} pairs for $\mathcal{D}^{\texttt{TM}}$.


\section{Data Preprocessing}

\subsubsection{Text Inpainting}

Previous works have shown that, in matching memes, removing text features is a crucial preprocessing step, where the space taken up by the text is usually blurred \cite{courtois_geneaology} or filled \cite{murgas_template}. This is due to text elements consisting of letters that comprise well-defined shapes and edges and that the same letters, and thus, same shapes and edges, are shared between all text. This could cause memes to be matched based on shared letters and fonts instead of key visual elements; artificially inflating similarity based only on shared text \cite{courtois_geneaology}. 
 
 To localise the portion of each image that contains text, we used EasyOCR\footnote{https://github.com/JaidedAI/EasyOCR}, with its \texttt{paragraph} mode, which returns contiguous text regions as opposed to a larger number of disconnected regions. These regions are set as the mask input for a DeepFillv2 
 model \cite{deepfill_v2}. DeepFill is a pretrained inpainting model that is trained to generate the pixel values within some masked region to minimise the visual discrepancy between the inpainted region and its surroundings. Courtois et al. used a similar inpainting step, where Gaussian noise was used instead of a pretrained inpainting model \cite{courtois_geneaology}. Whereas Dubey et al. finetuned a deep convolutional network to extract shared visual backgrounds while eliminating text overlays \cite{dubey_sparse}.

\subsubsection{Panel Segmentation}\label{sec:segment}
Memes often consist of multiple distinct visual segments \cite{hazman_what} that each express parts of a meme's entire meaning. Typically, some of these segments are, in themselves, the memetic element of a meme \cite{hazman_what}. For example, see Pair 4 in Table \ref{tab:memetic_matching}. Embedding- and hashing-based representations tend to favour similarity throughout the entire image (\textbf{whole-image} similarity) \cite{courtois_geneaology}. As such, we propose a segmentation approach that allows hashing- and embedding-based similarity to be computed between every segment in the reference image and each segment found within memes.

This segmentation step divides meme and reference images into (any) constituent segments, i.e. distinct panels akin to pages of a graphic novel. Here, we refer to \textit{segmentation} as the extraction of contiguous rectangular panels that each could constitute a standalone image; e.g., the two panels in Image A of Pair 4 in Table \ref{tab:memetic_matching}. This is in contrast to the more common sense of segmentation, which seeks to find meaningful areas or assigns class labels to individual pixels.

We use an open-source comic panel segmentation tool\footnote{\url{https://github.com/reidenong/ComicPanelSegmentation}}, which we chose for its low computational cost and its ability to exploit the clear horizontal and vertical lines common to comic pages and multimodal memes. This approach uses Canny edge detection followed by a Hough transform to split each image into panels. 

We found that this approach results in numerous segments that are \textit{blank} i.e., containing no discernible visual element.
Since these blank segments would spuriously match with one another in our segment‐wise similarity measures, we automatically filter them out by training a decision tree classifier. We used a dataset of 3,100 manually labelled segments (824 \texttt{blank}, 2,276 \texttt{non-blank}) which we randomly sampled from meme and reference images in both $\mathcal{D}^{\texttt{MM}}$ and $\mathcal{D}^{\texttt{TM}}$ KYM dictionary images and IMGFlip templates, and the meme datasets. The classifier uses Shannon entropy, aspect ratio, salience map entropy, and Laplacian variance as features. Although we initially included objectness scores and object counts from ObjectnessBING and a blankness ratio based on Otsu thresholding as input, these features were removed during Minimal Cost-Complexity Pruning (MCCP), reducing overfitting of the model. The optimal value for the Complexity parameter of MCCP, determined using a 10-fold cross-validation, resulted in the classifier achieving 87\% precision on the \texttt{non-blank} class.

A new instance of the classifier was then trained on all the data available from both the \texttt{MM} and \texttt{TM} datasets, using the tuned MCCP Complexity Parameter value. This was used to filter out blank segments on all $m$ and $r$ prior to computing segment-wise similarity measures.

\section{Evaluations}\label{method:similarity}
\subsection{Visual Similarity Measures}
 We compare the performance of six visual similarity measures on both the \texttt{MM} and \texttt{TM}: \texttt{Keypoint-D}, \texttt{Keypoint-M}, \texttt{Embed-W}, \texttt{Hash-W}, \texttt{Embed-S},  and \texttt{Hash-S}.

\subsubsection{\textbf{Keypoints}} To capture shared local features between memes, a keypoints-based approach was proposed in \cite{courtois_geneaology} which aims to match local features instead of whole-image similarity. To identify and represent visual keypoints in each image, we use \textit{Oriented FAST} and \textit{Rotated BRIEF}, respectively, similar to \cite{courtois_geneaology}. To measure the similarity between a meme image $m$ and a reference image $r$, we use OpenCV's \textit{BruteForceMatcher} to compare the descriptor of every keypoint in $m$ to that of every keypoint in $r$ using Hamming distance. For each keypoint in $m$, the distance of the closest match is returned. Thus, each pair $(m,r)$ is described by an array of distances $\Delta^{(m,r)}$.
In \cite{courtois_geneaology}, two memes are deemed related whenever the number of keypoint distances below a chosen maximum distance \(\delta^{\max}\) exceeds a minimum count \(C_{\min}\), i.e.
$
\sum_{i=1}^{|\Delta^{(m,r)}|} \mathbf{1}_{\{\Delta^{m,r,i} \le d_{\max}\}}
\;\ge\; C_{\min}.
$
Unlike in \cite{courtois_geneaology}, we use a more generalised representation of this criterion: cumulative frequency distribution of the distances in $\Delta^{(m,r)}$:
$$
F^{(m,r)} = \bigl|\{\delta \in \Delta^{(m,r)} : d \le x\}\bigr|,
\quad x = 0,\ldots,200,
$$
where $d$ is a distance value in \(\Delta^{(m,r)}\) and $x$ is limited to 200 as this was a greater value than any $\delta$ in all $(m,r)$ pairs in our datasets. This distribution constitutes our \emph{Keypoint Distance Distribution} (\textbf{\texttt{Keypoint-D}}) similarity representation. Our \emph{Keypoint Distance Moments} (\textbf{\texttt{Keypoint-M}}) representation consists of the mean, variance, skewness, and kurtosis of $F^{(m,r)}$.  

\par

\label{approach:embedding}\subsubsection{\textbf{Embeddings \& Hashing}} Embedding and hashing are two commonly used methods to create visual representations of images, where similar images create similar representations. Typically, embeddings are numerical arrays computed using a pretrained deep learning model, while hashing uses predefined transformations to represent images as binary codes. 

In this work, we use ResNeXt101-32x4d model pretrained on ImageNet1k and Perceptual Hashing (pHash) to represent each image in our embeddings- and hashing-based similarity measures, respectively. The similarity between two images are computed as the cosine distance between two embeddings or the Hamming distance between two hashes. Based on these, we define four similarity measures based on: Whole Image Embedding (\texttt{Embed-W}), Whole-Image Hashing (\texttt{Hash-W}), Segment-wise Embedding (\texttt{Embed-S}), and Segment-wise Hashing (\texttt{Hash-S}).




\textbf{Whole-Image Similarity}: 
We represent the Whole Image similarity measures, \textbf{\texttt{Embed-W}} and \textbf{\texttt{Hash-W}}, of a given pair \((m,r)\) as the whole‐image similarity of $r$ and $m$, $\sigma_w^{(m,r)}$, and 
the similarity rank, $\mathrm{rank}_w(m,r)$. The former is calculated as:
\[
\sigma_w^{(m,r)} = \mathrm{dist}\bigl(f(m),\,f(r)\bigr)
\]
where \(f(\cdot)\) is the embedding of an image and $\mathrm{dist}$ is the cosine distance for \texttt{Embed-W}. Whereas in \texttt{Hash-W}, \(f(\cdot)\) is the pHash of an image and $\mathrm{dist}$ is the Hamming distance. We define $\mathcal{V}_w^{(m,R)}$ as $\{\sigma_w(m,r') \mid r'\in R\}$ for any $m$ and all $r'$ in $R$.

We use a ranking operator, which returns the rank of any $\sigma$ in $\mathcal{V}$ in descending order. The similarity rank of $(m,r)$ relative to the similarity of $m$ to $r' \in R$ is calculated as $$\mathrm{rank}_w(m,r) = \mathrm{rank}\bigl(\sigma_w^{(m,r)};\,\mathcal{V}_w^{(m,R)}\bigr)$$

\textbf{Segment-Wise Similarity}: Noting that \texttt{related} pairs in $\mathcal{D}^{\texttt{MM}}$ require at least one similar segment (e.g., Pair 4 in Tab. \ref{tab:memetic_matching}), while \texttt{TM} annotations typically require all segments to be similar (e.g. Pair 2 in Tab. \ref{tab:memetic_matching}). As such, we propose a segment-wise version of embedding-based and hashing-based similairity measures, \textbf{\texttt{Embed-S}} and \textbf{\texttt{Hash-S}} respectively. 

Given that $m$ and $r$ comprise segments $S^m=\{s_i^m : i=1,\dots,n_m\}$ and $S^r=\{s_j^r : j=1,\dots,n_r\}$, respectively, where $n_m$ and $n_r$ are the numbers of non-blank segments in $m$ and $r$. We represent each pair $(m,r)$ by the mean, minimum, maximum, and spread of segment-wise similarity scores $\mathcal{V}_s^{(m,r)}$, the minimum and maximum ranks of these scores relative to all segments in $\mathrm{rank}(\mathcal{V}_s^{(m,r)};\,\mathcal{V}_s^{(m,R)})$, and the number of segments found in each image,  $n_m$ and $n_r$.

We compute the similarity between every segment in $m$ and every segment in $r$ as:
$$ \mathcal{V}_s^{(m,r)} = \{\sigma_s(s_i^m,s_j^r) \mid \ s_i^m\in S(m),\ s_j^r\in S(r) \}
$$

To allow us to rank the similarity of $(m,r)$ relative to all reference images in $R$, we also collect similarity scores between every segment in $m$ and every segment in all reference images, $R$, in
$\mathcal{V}_s^{(m,R)}= \{\mathcal{V}_s^{(m,r')} \mid r' \in R\}$. For each pair $(m,r)$, we compute the maximum and minimum of $\mathrm{rank}(x;\,\mathcal{V}_s^{(m,R)})$, respectively, where $x \in \mathcal{V}_s^{(m,r)}$.



\begin{figure*}[t]
    \centering
    \includegraphics[width=1\linewidth]{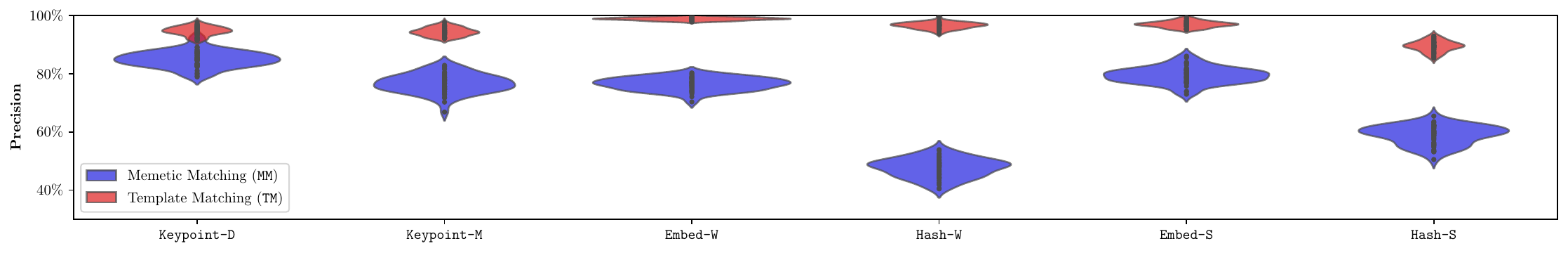}
    \caption{The Precision score distribution over 50 train-test splits of 6 similarity measures on the \texttt{related} class as evaluated on the \texttt{MM} and \texttt{TM} tasks.
    }
    \label{fig:results}
\end{figure*}

\subsection{Similarity-based Classifier}



 For each task $t$, we generate 50 random train-test splits $\{(\mathcal{D}_{split}^{(t,j)} \mid j = 1,\ldots,50 \}$ where $split \in \{\text{train},\text{test}\}$. For each split $j$, similarity measure $s$, and task $t$, we compute the feature vectors using the representations described above: 
\begin{align*}
X_{split}^{(t,s,j)},\ \hat{y}_{\text{}}^{(t,j)} &= {\left(s(m, r),\ \hat{y}\right) \mid (m, r, \hat{y}) \in \mathcal{D}_{split}^{(t,j)} }
\end{align*}

The performance distribution of a given $s$ on $t$ is given by the performance of a Decision Tree classifier, $C^{(t,s,j)}$. For each $j$, we trained an instance in each $(X_{\text{train}}^{(t,s,j)}, \hat{y}_{\text{train}}^{(t,j)})$ and evaluated its predictions on $X_{\text{test}}^{(t,s,j)}$ against $\hat{y}^{(t,j)}_{\text{test}}$.


With the resulting performance distributions, we assess: (1) the relative classification effectiveness of each similarity measure, and (2) how the performance ranking of methods varies between the \texttt{MM} and \texttt{TM} tasks. For our performance metric, we use
the precision on the \texttt{related} class across the 50 splits. This indicates how well $C^{(t,s,j)}$ minimises false positives. A low precision implies that unrelated image pairs are being misclassified as matches, which would lead to erroneous matches polluting downstream analyses and irrelevant dictionary lookup.

\subsection{MLLM as a Meme Matcher} 
To tackle \texttt{MM} by prompting a pretrained MLLM requires us to select an MLLM model and a prompt with which to invoke the model to respond with either \texttt{related} or \texttt{unrelated} on each given $(m,r)$. We selected the open-source LLaVA-OneVision with 7 billion parameters as our MLLM. To craft our prompt, we applied the Instruction Induction method \cite{honovich-instruction_induction}. We use this method to generate 16 candidate prompts, each using 8 random samples of $(m,r)$ pairs from $\mathcal{D}^{\texttt{MM}}$. Each candidate prompt was evaluated on $\mathcal{D}_{test}^{(t=\texttt{MM},j)}$, for $j = 1,\ldots,50$. The prompt achieving the highest mean precision was selected for downstream evaluation.

We evaluated the MLLM using this prompt in three settings: (1) \texttt{Zero-Shot} uses the prompt as it was generated, requesting the MLLM to output either \texttt{related} or \texttt{unrelated}. (2) \texttt{Zero-Shot CoT} appends to the prompt a request for the MLLM to generate justifications prior to its predicted classification output. (3) \texttt{Few-Shot} uses the same prompt as \texttt{Zero-Shot CoT} but includes two in-context demonstrations. For each input pair $(m,r)$, we assign two other labelled pairs , $(m',r')$, from $\mathcal{D}^{\texttt{MM}}$, one for each class, as demonstrations. In choosing $(m',r')$, if available, we use a pair where $r' = r$ but $m' \not= m$. If no such pair exists, we use pairs where $r'$ is most similar to $r$ and $m' \not= m$. If a chosen $r'$ is included in multiple $(m',r')$ pairs, we select the pair where $m'$ is most similar to but not the same as $m$. Here, similarity is measured using the cosine distance of the embeddings as is used in \texttt{Embed-W}.


\section{Results}

\begin{figure}[b]
    \centering
    \includegraphics[width=0.8\linewidth]{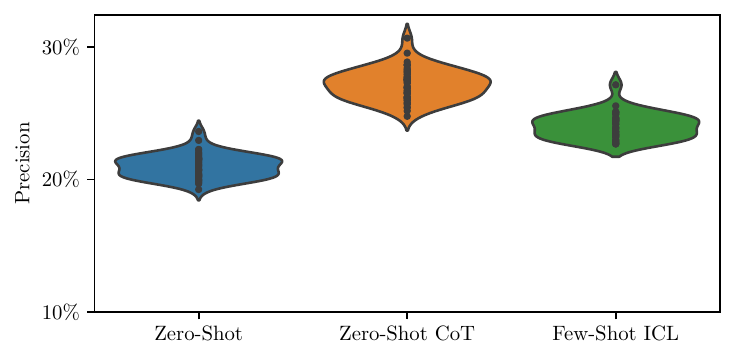}
    \caption{Precision score distribution of 3 MLLM-based methods on \texttt{MM}.}
    \label{fig:llmresults}
\end{figure}

The performance distribution of the six similarity measures, as shown in Fig. \ref{fig:results}, shows a clear distinction between the \texttt{MM} and \texttt{TM} tasks. The precision scores were consistently lower for \texttt{MM} than \texttt{TM}. This difference was found to be statistically significant when evaluated using the Mann-Whitney U test. Between the two tasks, on each similarity measure, the p-values were found to be 1.36e-17, 7.06e-18, 7.05e-18, 7.03e-18, 7.05e-18 and 7.06e-18 for \texttt{Keypoint-D} , \texttt{Keypoint-M}, \texttt{Embed-W}, \texttt{Hash-W}, \texttt{Embed-S}, and \texttt{Hash-S}, respectively.  These results indicate that \texttt{MM} is the more computationally challenging of the two tasks. 

These results also show that the most suitable similarity measure depends on the formulation of meme matching used. Specifically, \texttt{Keypoint-D} performed best in 48 out of 50 splits for the task of \texttt{MM} while \texttt{Embed-W} did so in 49 splits in the \texttt{TM} task. Thus, the most suitable similarity measure for \texttt{TM}, e.g., as put forward in \cite{murgas_template}, should not be assumed to be the best choice for \texttt{MM}.
 The strong performance of \texttt{Embed-W} and \texttt{Hash-W} for \texttt{TM} aligns with the findings presented by Murgas et al.. They reported a weak performance of keypoint similarity measures relative to embedding-based methods. Our analysis suggests that their observation stems from their choice to focus solely on template-based memes and to ignore other memetic elements when annotating their data.

For \texttt{MM}, segment-wise methods, \texttt{Embed-S} and \texttt{Hash-S}, significantly outperformed their whole-image counterparts, with Wilcoxon test p-values of 8.97e-9 and 1.77e-15, for embedding- and hashing-based approaches, respectively. This supports the arguments made by \cite{courtois_geneaology}, who emphasised the need to capture local memetic elements instead of just whole-image similarity. Our results show that, of the similarity measures evaluated here, \texttt{Keypoint-D} performed well on both \texttt{MM} and \texttt{TM}, making it a viable meme matching method for memetics- and template-based meme analysis tasks.


Surprisingly, we found that the MLLM-based matching methods underperformed all other methods on the \texttt{MM} task, as shown in Fig. \ref{fig:llmresults}. This may stem from several issues: First, as is the case in almost all prompting-based solutions, the performance is highly dependent on the model used. The low performance observed may be due to the size of the chosen model -- the one tested here has only 7B parameters, while GPT-4 is estimated to have 1.7T. Second, prompt engineering could be used to further improve task performance for any given model. Since our prompts are written by the MLLM, our results can be further attributed to the model size chosen; as smaller models have been shown to struggle with editing and writing prompts on certain tasks \cite{min-etal-2022-rethinking}. Finally, it is unclear how similar the \texttt{MM} and \texttt{TM} tasks are to the tasks used when pretraining the MLLM. Considering all these factors, a potential future direction for creating MLLM-based meme matching solutions could involve fine-tuning a larger model on a sufficiently large collection of manually annotated matches.

\section{Conclusions}
In this work, we proposed the Memetic Matching task as a more realistic and comprehensive approach to meme matching than the current approaches of Template Matching. Our evaluation found that visual similarity measures commonly used on the meme matching task significantly underperform when using a memetic-based definition of memes, which contrasts with the results found when only using a template-based definition of memes.
For Memetic Matching, we showed that a keypoint-based approach outperformed both embedding- and hash-based approaches. Furthermore, we found that our novel segment-wise conmputation of embedding and hash similarity outperformed similarity measures based on whole-image visual representations. Finally, we found that a mid-sized open-source general purpose MLLM does not yet pose a competitive solution to Memetic Matching; motivating more in-depth analyses and broader design approaches towards an MLLM-based solution to the task. Taken together, our observations suggest that further research is needed to enable reliable retrieval of information via memetic elements other than just meme templates. 

\printbibliography

\end{document}